# EXTENDING DEPTH OF FIELD FOR VARIFOCAL MULTIVIEW IMAGES

*Zhilong Li, Kejun Wu, Qiong Liu, and You Yang*

School of Electronic Information and Communications,
Huazhong University of Science and Technology, China

## ABSTRACT

Optical imaging systems are generally limited by the depth of field because of the nature of the optics. Therefore, extending depth of field (EDoF) is a fundamental task for meeting the requirements of emerging visual applications. To solve this task, the common practice is using multi-focus images from a single viewpoint. This method can obtain acceptable quality of EDoF under the condition of fixed field of view, but it is only applicable to static scenes and the field of view is limited and fixed. An emerging data type, varifocal multiview images have the potential to become a new paradigm for solving the EDoF, because the data contains more field of view information than multi-focus images. To realize EDoF of varifocal multiview images, we propose an end-to-end method for the EDoF, including image alignment, image optimization and image fusion. Experimental results demonstrate the efficiency of the proposed method.

***Index Terms—*** Varifocal multiview images, Extending depth of field, Image alignment, Image fusion

## 1. INTRODUCTION

Extending depth of field (EDoF) is a key function for many visual applications, such as microscopic imaging, endoscopy, and even visual products in the field of consumer electronics [1-3]. For the traditional visual EDoF, it is mainly aimed at the static scene, and EDoF is realized by imaging the static scene with multiple different focal lengths [4]. This method is more conducive to obtaining high-quality and high-signal-to-noise ratio focal lengths images [5]. It is clear that the traditional method is to obtain multi-focus images at different times, which will not be conducive to EDoF of dynamic scenes. And the field of view of the imaging will be limited to a single view, which is not conducive to the observation of more extensive scene information. The key point to achieve EDoF of static scene is to obtain multiple images with different focal lengths. Therefore, to expand the field of view and facilitate EDoF of dynamic scenes, the key point is to obtain multi-focus images at the same time.

This work is supported by National Natural Science Foundation of China (No. 61971203).

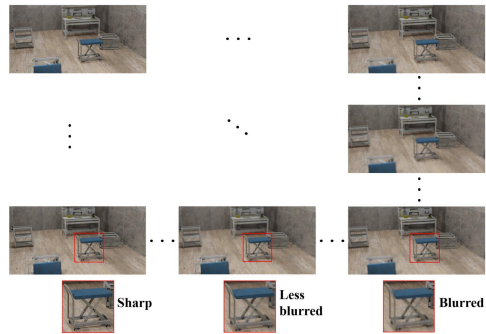

**Fig. 1**. Varifocal multiview image data type.

For solving different visual problems, multi-view method is a common idea. For example, multi-view can solve the problem of different fields of view, obtain more realistic and natural stereoscopic images, and deal with multi-spectral image reconstruction [6,7]. According to the multi-view method, Li and Chowdhury et al. proposed using multi-focus multi-view data type for 3D scene reconstruction [8,9]. For this emerging varifocal multiview data type, each source image comes from different perspectives and focuses on different depths. Compared with single-view multi-focus images, varifocal multiview images have a larger field of view, and also have multi-focus data types, which are conducive to EDoF. According to the characteristics of this type of images, this paper uses the varifocal multiview image dataset of Ornament Scene and Furniture Scene [10] to EDoF. In Ornament Scene and Furniture Scene, each view is focused at different depths, so we can obtain images with different focal lengths at the same time by imaging the same scene from multiple viewpoints, so as to obtain the basic elements of EDoF. Meanwhile, since the images of each viewpoint are obtained at the same time, we can use this data type to record the dynamic scene repeatedly at different times. However, for varifocal multiview data types, each image comes from different viewpoint and different focus depths, so there are location differences between images. To solve this problem, we propose a EDoF scheme. Firstly, we determine the benchmark view of EDoF, and transform the images from other viewpoints to the benchmark view by perspective transformation method. To ensure the integrity of the transformation information, the homography matrix is modified. Subsequently, to obtain the sharpest image at the corresponding focus and reduce data redundancy, source images are split into blocks with fixed size, and the optimal sub-images are selected for fusion to obtain the result of EDoF. Finally, fusion results are spliced to obtain the final scene image after EDoF. Our contributions of this work can be concluded as follows:

- For the novel representation of varifocal multiview data types, an end-to-end methods are proposed. Based on this method, the scene images focusing on different depths can be clearer to achieve the purpose of EDoF, and even the effect of full EDoF can be achieved to a certain extent. This can contribute to future vision systems and applications to handle the challenges of EDoF of varifocal multiview image data types, and will also facilitate EDoF research in dynamic scenes.

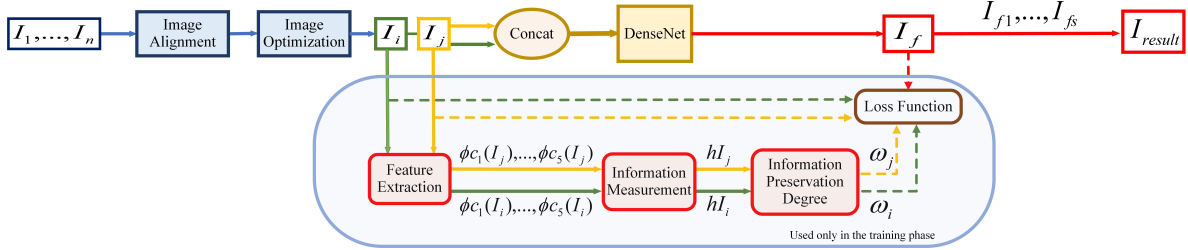

**Fig. 2**. Pipeline of processing method proposed in this paper. Dashed lines represent the data used in the loss function.

## 2. THE PROPOSED METHOD

For the varifocal multiview image data types, as shown in Fig. 1. Each image is an independent individual from different perspectives, and the images from adjacent perspectives focus on different depths. Therefore, for varifocal multiview images, the sharpness of the same position will show a clearer or more blurred trend with the change of focus position. In addition, since each viewpoint contains different scene information, the use of multi-view images for EDoF will help to obtain larger field of view results. However, since each image comes from a different perspective and focuses on different depths, there are horizontal and vertical parallaxes and focus position inconsistency between images.

According to the characteristics of varifocal multiview image data types, the proposed processing method is shown in Fig. 2. With source images denoted as $I_1, \ldots, I_n$, due to the horizontal and vertical parallaxes and focus position inconsistency between images, the source images are aligned firstly, and the images of each viewpoint are transformed into the selected benchmark view. Secondly, the aligned images are split, and the sharpest sub-images $I_i$ and $I_j$ are selected, a DenseNet is trained to generate the fusion image $I_f$. The outputs of feature extraction are the feature maps $\phi c_1(I_i), \ldots, \phi c_5(I_i)$ and $\phi c_1(I_j), \ldots, \phi c_5(I_j)$. Then the information measurement is performed on these feature maps, producing two measurements denoted by $hI_i$ and $hI_j$. With subsequent processing, the information preservation degrees are denoted as $\omega_i$ and $\omega_j$. $I_i$, $I_j$, $I_f$, $\omega_i$ and $\omega_j$ are used in the loss function without the need for ground truth. In the training phase, $\omega_i$ and $\omega_j$ are measured and applied in defining the loss function. Then, a DenseNet module is optimized to minimize the loss function. In the testing phase, $\omega_i$ and $\omega_j$ do not need to be measured, as the DenseNet has been optimized. Finally, the final result image $I_{result}$ is obtained by splicing the sub-image $I_{f1}, \ldots, I_{fs}$ of each position after fusion. The detailed definitions or descriptions are given in the following subsections.

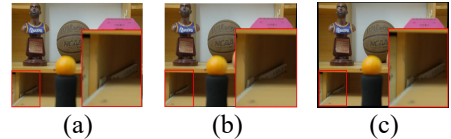

(a)          (b)          (c)

**Fig. 3**. The results of image alignment. (a) Source image. (b) The result of perspective transformation. (c) The result of the modified homography matrix.

### 2.1. Image Alignment

Due to the position deviation between the source images captured, it is necessary to align the images before EDoF. We first use SURF [11] feature point detection algorithm to extract the effective feature points in each image, and match the feature points under the selected benchmark view to find the most matching feature point pair. Subsequently, extracting the coordinates of the optimal matching feature point pair to calculate the homography matrix, and perform perspective transformation on the images under each viewpoint to generate the aligned images. After the perspective matrix transformation, the position coordinates of the source image in the new coordinate system will become negative, resulting in the partial information being intercepted directly. To preserve the complete scene information, the homography matrix is modified so that the aligned images has a deviation in the horizontal and vertical directions to restore the lost scene information. The results of image alignment are shown in Fig. 3.

### 2.2. Image Optimization

After image alignment, to reduce the data redundancy and obtain the sharpest image at the corresponding focus position, we split the aligned image into nine sub-images with the same size, analyze the required focus depth of each sub-image, and quantitatively analyze each sub-image using the mean gradient metrics [12]. Usually, the greater the mean gradient is, the more the details of the image are, and the higher the sharpness is. After selecting the sharpest sub-images, the image fusion is performed to obtain the result of the corresponding position EDoF.

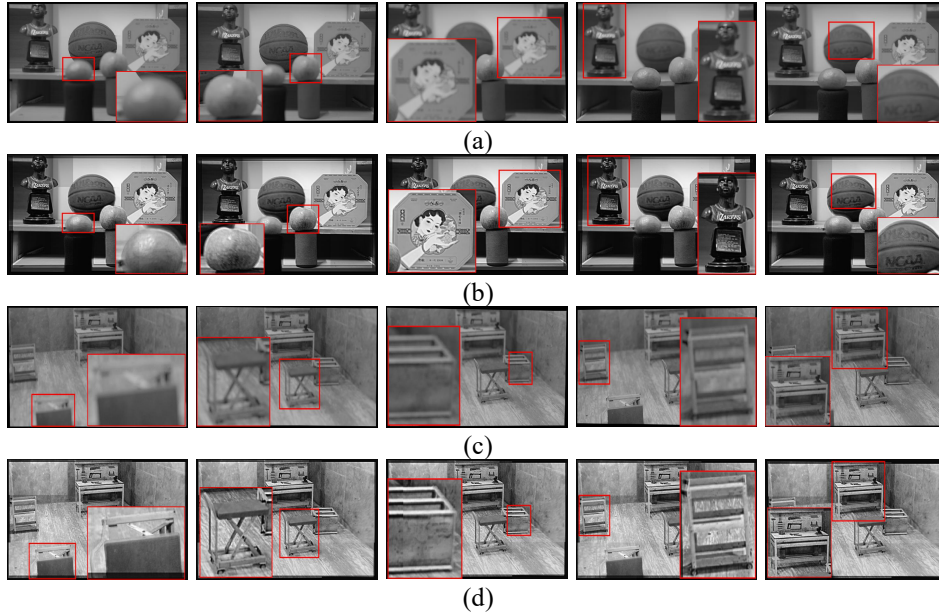

**Fig. 4**. Qualitative comparison between EDoF images and source images. (a) Source images of Ornament Scene. (b) EDoF images of Ornament Scene. (c) Source images of Furniture Scene. (d) EDoF images of Furniture Scene.

### 2.3. Feature Extraction

In other computer vision tasks, larger and more diversified datasets are often used to train models compared to image fusion tasks. Thus, features extracted by such models are abundant and comprehensive [13,14]. we adopt the pretrained VGG-16 network [15] for feature extraction to extract both shallow-level features (textures, local shapes) and deeplevel features (content, spatial structures) for estimating the information measurement.

### 2.4. Information Measurement

To measure the information contained in the extracted feature maps, the image gradients are used to evaluate. When gradient is used in the deep learning framework, it will be better for computation and storage. Thus, they are more suitable for application in CNN for information measurement. The information measurement is defined as follows:

$$hI = \frac{1}{5}\sum_{p=1}^{5} \frac{1}{H_p W_p D_p} \sum_{k=1}^{D_p} \left\| \nabla \phi c_p^k(I) \right\|_F^2, \quad (1)$$

where $\phi c_p(I)$ is the feature map by the convolutional layer before the $p$-th max-pooling layer. $k$ denotes the feature map in the $k$-th channel of $D_p$ channels. $\|\cdot\|_F$ denotes the Frobenius norm, and $\nabla$ is the Laplacian operator.

### 2.5. Information Preservation Degree

To preserve valid information in source images, two adaptive weights are designed as information preservation degrees, which define the weights of similarities between the fusion image and the source images. The higher the weight, the higher the information preservation degree of the corresponding source image is. These adaptive weights, denoted as $\omega_i$ and $\omega_j$, are estimated according to the information measurement results $hI_i$ and $hI_j$ obtained by Eq. (1). Thus, $\omega_i$ and $\omega_j$ are defined as:

$$[\omega_i, \omega_j] = \text{softmax}([\tfrac{hI_i}{c}, \tfrac{hI_j}{c}]), \quad (2)$$

where we use the softmax function to map $\tfrac{hI_i}{c}, \tfrac{hI_j}{c}$ to real numbers between 0 and 1, and guarantee that the sum of $\omega_i$ and $\omega_j$ is 1.

### 2.6. Loss Function

The loss function is mainly used to preserve important information and to train a single model defined as follows:

$$\mathcal{L}(\theta, D) = \mathcal{L}_{\text{sim}}(\theta, D), \quad (3)$$

where $\theta$ denotes the parameters in DenseNet, and $D$ is the training datasets. $\mathcal{L}_{\text{sim}}(\theta, D)$ is the similarity loss between the result and source images. The similarity constraint come from structural similarity and intensity distribution [16]. We use the structural similarity index measure (SSIM) [17] to constrain the structural similarity between $I_i$, $I_j$, and $I_f$.

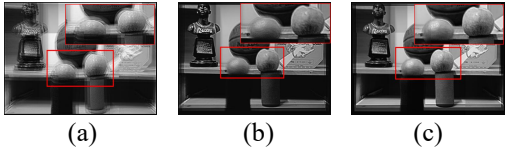

(a)　　　　　　(b)　　　　　　(c)

**Fig. 5**. Qualitative results of ablation experiment on image alignment and image optimization. (a) Result without image alignment. (b) Result without image optimization. (c) Result with image alignment and optimization.

Thus, with $\omega_i$ and $\omega_j$ to control the information degree, the first item of $\mathcal{L}_{\text{sim}}(\theta, D)$ is formulated as:

$$\mathcal{L}_{\text{ssim}}(\theta, D) = \mathrm{E}[\omega_i \cdot (1 - S_{I_f, I_i}) + \omega_j \cdot (1 - S_{I_f, I_j})], \quad (4)$$

where $S_{x,y}$ denotes the SSIM value between two images.

While SSIM focuses on the changes of contrast and structure, it shows weaker constraints on the difference of the intensity distribution. We supplement $\mathcal{L}_{\text{ssim}}(\theta, D)$ with the second item, which is defined by the mean square error (MSE) between two images:

$$\mathcal{L}_{\text{mse}}(\theta, D) = \mathrm{E}[\omega_i \cdot \mathrm{MSE}_{I_f, I_i} + \omega_j \cdot \mathrm{MSE}_{I_f, I_j}], \quad (5)$$

with $\alpha$ controlling the trade-off, $\mathcal{L}_{\text{sim}}(\theta, D)$ is formulated as:

$$\mathcal{L}_{\text{sim}}(\theta, D) = \mathcal{L}_{\text{ssim}}(\theta, D) + \alpha \mathcal{L}_{\text{mse}}(\theta, D). \quad (6)$$

## 3. EXPERIMENTS AND DISCUSSIONS

### 3.1. Experiment Setting

We aim to extend the depth of field of varifocal multiview images. Two test sequences Ornament Scene and Furniture Scene selected from [10] are employed in our experiments. Subsequently, we extract the varifocal multiview image of 3x3 size from test sequence as the source images. Each source image comes from different perspectives, and the source images of adjacent perspectives are focused at different depths to ensure that all perspectives are fully focused on each location of the scene, which is taken as shooting the scene at the same time.

### 3.2. Experimental Results

The experimental results of EDoF of the source images of Ornament Scene and Furniture Scene using the method proposed in this paper are shown in Fig. 4. In the figure, the first line and the third line are the source images of Ornament Scene and Furniture Scene, respectively. The second line and the fourth line are the results of EDoF in each scene, respectively. Compared with the source images at each focusing depth, the method proposed in this paper can make the images at each depth clearer, achieve the purpose of EDoF, and realize the effect of full EDoF to a certain extent. Besides, the multi-view method is used to solve the multi-focus problem, which can enlarge the field of view of the resulting image after EDoF, and help to observe the richer scene information under the benchmark view.

**Table 1**. Quantitative results of ablation experiment on image alignment and image optimization. Bold numbers indicate the optimal value for each set of data.

| Metrics | Without alignment | Without optimization | With alignment and optimization |
|---|---|---|---|
| IE | 6.8152 | 5.8158 | **6.8993** |
| LC | 0.8932 | 0.8578 | **0.8986** |

### 3.3. Ablation Experiments

For varifocal multiview images, there are horizontal and vertical parallaxes and focus position inconsistency among images. In order to ensure EDoF effect, we first perform image alignment. Moreover, in order to reduce data redundancy and obtain the sharpest image at the corresponding focus position, we also optimize the image processing. To verify the effectiveness of the two processing modules, we conducted ablation experiments. Firstly, we carried out a comparative experiment without image alignment, and compared the results without image alignment with the results after alignment. The qualitative results, as shown in Fig. 5 (a) and (c), show that the source image without image alignment can lead to serious artifacts and can not accurately determine the scene information, which is not conducive to the understanding of the scene. Then, we also carried out a comparative experiment without image optimization. The qualitative results are shown in Fig. 5 (b) and (c). It can be seen from the figure that the optimization module processing of the source images will enhance the clarity of the image, and also alleviate the artifacts of the image, so that the effect of EDoF is better. In addition, we also use two objective evaluation metrics of information entropy (IE) [18] and local contrast (LC) [19] for quantitative analysis. Higher IE and LC show that the image contains more details. The better contrast of the image, the better the visual effect of EDoF. The calculation results of the two metrics are shown in Table 1. The results of image alignment and optimization have achieved the optimal results in IE and LC, indicating that image alignment and image optimization can improve the clarity of the image, enhance the image contrast, and make the visual effect of EDoF better.

## 4. CONCLUSION

In this paper, an end-to-end method is proposed to extend the depth of field of the emerging varifocal multiview images. And the varifocal multiview images of Ornament Scene and Furniture Scene are used as experimental data to carry out EDoF experiment. The experimental results show that the method can achieve better EDoF for varifocal multiview images. This will be conducive to future visual systems and applications to handle the challenges of EDoF of varifocal multiview images, and even in the dynamic scene EDoF research.